\documentclass[runningheads]{llncs}
\usepackage{graphicx}
\usepackage{tikz}
\usepackage{comment} 
\usepackage{amsmath,amssymb} % define this before the line numbering.
\usepackage{color}
\usepackage{adjustbox}

\usepackage[ruled,vlined]{algorithm2e}

\usepackage{enumitem}
\usepackage{multirow}

\begin{document}
% \linenumbers
\pagestyle{headings}
\mainmatter
\def\ECCVSubNumber{5752}  % Insert your submission number here

\title{Learning to Transfer Learn: Reinforcement Learning-Based Selection for Adaptive \\ Transfer Learning} % Replace with your title

\begin{comment}
\titlerunning{ECCV-20 submission ID \ECCVSubNumber} 
\authorrunning{ECCV-20 submission ID \ECCVSubNumber} 
\author{Anonymous ECCV submission}
\institute{Paper ID \ECCVSubNumber}
\end{comment}
%******************

%\begin{comment}
\titlerunning{Learning to Transfer Learn}
% First Author\inst{1}\orcidID{0000-1111-2222-3333} \and
% Third Author\inst{3}\orcidID{2222--3333-4444-5555}
\author{
 Linchao Zhu\inst{1,2}
 \and
 Sercan \"{O}. Ar{\i}k\inst{1}
 \and
 Yi Yang\inst{2}
 \and
 Tomas Pfister\inst{1}
}
\authorrunning{Zhu et al.}
\institute{Google Cloud AI, Sunnyvale, CA  \\
\and
 University of Technology Sydney, Sydney, Australia \\
\email{\{soarik,tpfister\}@google.com};
\email{\{linchao.zhu,yi.yang\}@uts.edu.au}
}
%\end{comment}
%******************
\maketitle

\begin{abstract}
We propose a novel adaptive transfer learning framework, learning to transfer learn (L2TL), to improve performance on a target dataset by careful extraction of the related information from a source dataset. Our framework considers cooperative optimization of shared weights between models for source and target tasks, and adjusts the constituent loss weights adaptively. The adaptation of the weights is based on a reinforcement learning (RL) selection policy, guided with a performance metric on the target validation set. We demonstrate that L2TL outperforms fine-tuning baselines and other adaptive transfer learning methods on eight datasets. In the regimes of small-scale target datasets and significant label mismatch between source and target datasets, L2TL shows particularly large benefits.

\end{abstract}

\section{Introduction}
Deep neural networks excel at understanding images \cite{resnet,vdcnn_image}, text \cite{bert} and audio \cite{wavenet,deepspeech2}.
The performance of deep neural networks improves significantly with more training data \cite{DL_scaling}. 
As the applications diversify and span use cases with small training datasets, conventional training approaches are often insufficient to yield high performance. It becomes highly beneficial to utilize extra source datasets and ``transfer" the relevant information to the target dataset. Transfer learning, commonly in the form of obtaining a pre-trained model on a large-scale source dataset and then further training it on the target dataset (known as fine-tuning), has become the standard recipe for most real-world artificial intelligence applications. Compared to training from random initialization, fine-tuning yields considerable performance improvements and convergence speedup, as demonstrated for object recognition \cite{cnn_shelf}, semantic segmentation \cite{cnn_semantic_segmentation}, language understanding \cite{bert}, speech synthesis \cite{voice_cloning}, audio-visual recognition \cite{audio_visual} and language translation \cite{machine_translation}. 

\begin{figure}[t]
\centering
\includegraphics[width=1.0\linewidth]{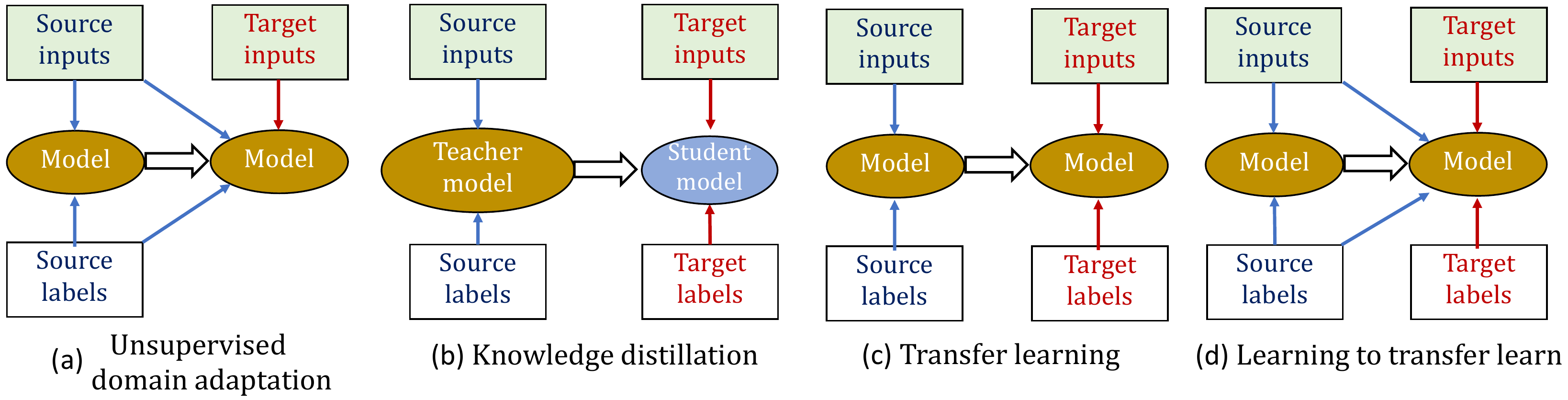}
\caption{{L2TL and other adaptation settings}.
(a) Unsupervised domain adaptation incorporates source data and source labels for target domain adaptation, where the target labels are not provided.
(b) Knowledge distillation aims to distill source knowledge from a teacher model to a student model, and the student model is usually more lightweight.
(c) Conventional transfer learning transfers knowledge via weights of the model pre-trained on a source dataset, to obtain better performance on the target dataset.
(d) Our L2TL adaptively infers the importance weights of the source examples based on the feedback from the target objective on the target validation set. With adaptive assignment, more relevant examples get higher weights to extract the information from the source dataset.
}
\label{fig:intro}
\end{figure}

Towards the motivation of pushing the performance of transfer learning, recent studies \cite{ngiam2018domain,weakly_supervised_TL,biobert,liu2019understanding} have explored the direction of matching the source and target dataset distributions. Even simple methods to encourage domain similarity, such as prior class distribution matching in Domain Adaptive Transfer Learning (DATL)~\cite{ngiam2018domain}, are shown to be effective -- indeed, in some cases, more important than the scale of the source dataset. Such adaptive transfer learning approaches, as in L2TL, typically assume the availability of the labeled source dataset for training on the labeled target dataset (that also differentiates the setting from unsupervised domain adaptation \cite{gong2012geodesic} or knowledge distillation \cite{hinton2015distilling}, see Fig.~\ref{fig:intro}), along with the pre-trained model.
Given the increasing availability of very-large scale public datasets for various data types and the demand for cutting-edge deep learning on highly-specialized target tasks with small training datasets, this setting is indeed getting very common in practice \cite{cui2018large,ge2017borrowing,ngiam2018domain}.

In this paper, our goal is to push this direction further by introducing a novel \textit{reinforcement learning (RL)-based framework}. Our framework, learning to transfer learn (L2TL), adaptively infers the beneficial source samples directly from the performance on the target task. 
There are cases that source samples could have features that are implicitly relevant to the target samples and would benefit the learning process, but they may belong to different classes. For example, consider the classification problem for bird images. The source dataset may not contain bird images, but may have airplane images with similar visual patterns that would aid the training of the bird classifier as they share similar visual patterns to learn valuable representations of the raw data. 
\textit{L2TL framework is designed to automatically handle such cases with its policy learning, and can push the performance further in ways that manual source dataset selection or fixed domain similarity methods may not be able to}. 
L2TL considers cooperative optimization of models for source and target tasks, while using adaptive weights for scaling of constituent loss terms. 
L2TL leverages the performance metric on the target validation set as the reward to train the policy model, which outputs the weights for each source class adaptively. Overall, L2TL does not utilize an explicit similarity metric as in \cite{cui2018large,ngiam2018domain}, but learns source class weights to directly optimize the target dataset performance.

We demonstrate promising transfer learning results given fixed models in a wide range of scenarios:
\begin{itemize}[leftmargin=*, noitemsep]
\item \emph{Source and target datasets from similar domains}: L2TL consistently outperforms the fine-tuning baseline with a 0.6\%-1.3\% relative accuracy gain on five fine-grained datasets, and DATL \cite{ngiam2018domain} with a 0.3\%-1.5\% relative accuracy gain. When the similarities become very apparent between some source and target classes, e.g. as in MNIST to SVHN digit recognition transfer case, the relative accuracy gain is {3.5\%} compared to fine-tuning baseline. 
\item \emph{Large-scale source datasets}: When very large-scale source datasets are used, the selection of the relevant source classes become more important, the gain of L2TL is up to \textbf{7.5\%}.
\item \emph{Low-shot target dataset regime}: L2TL significantly outperforms fine-tuning on fine-grained target datasets, up to \textbf{6.5\%} accuracy gain with five samples per class.
\item \emph{Source and target datasets from dissimilar domains}: While other advanced transfer learning (based on explicit similarity measures) cannot be readily applied for this scenario, L2TL outperforms the fine-tuning baseline, up to \textbf{1.7\%} accuracy gain on a texture dataset and \textbf{0.7 AUC} gain on Chest X-Ray dataset. % \cite{irvin2019chexpert}.
\end{itemize}
In addition, L2TL yields ranking of the source data samples according to their contributions to the target task, that can open horizons for new forms of interpretable insights.

\section{Related Work}
\noindent\textbf{Adaptive transfer learning:} There is a long history of transfer learning for neural networks, particularly in the form of fine-tuning \cite{rich_feature_hiar}. Various directions were recently considered to improve standard fine-tuning. One direction is carefully choosing which portion of the network to adapt while optimizing the information extraction from the source dataset. In \cite{spottune}, a policy network is used to make routing decisions on whether to pass the input through the fine-tuned or the pre-trained layers. In \cite{inductive_bias_TL}, a regularization scheme is proposed to promote the similarity of the fine-tuned model with the pre-trained model as a favorable inductive bias. Another direction is carefully choosing which input samples are relevant to the target task, as in our paper. 
\cite{ge2017borrowing} uses filter bank responses to select nearest neighbor source samples and demonstrates improved performance. In \cite{cui2018large}, domain similarity between source and target datasets is quantified using Earth Mover’s Distance (EMD). Transfer learning is shown to benefit from pre-training on a source domain that is similar in EMD. With a simple greedy subset creation selection criteria, promising results are shown for improving the target test set performance. Domain adaptive transfer learning (DATL) \cite{ngiam2018domain} employs probabilistic shaping, where the value is proportional to the ratios of estimated label prior probabilities. L2TL does not use a similarity metric like proximity of filter bank responses, EMD or prior class probabilities. Instead, it aims to assign weights to optimize the target set metric directly. 

\noindent\textbf{Reweighing training examples:} Reweighing of constituent training terms has been considered for various performance goals.
\cite{ren2018learning} applies gradient descent-based meta-learning to update the weights of the input data, with the goal of providing more noise-robust and class-balanced learning.
\cite{jenni2018deep} formulates reweighing as a bilevel optimization problem, such that higher weights are encouraged for the training samples with more agreement of the gradients on the validation set.
Focal loss \cite{focal_loss} is another soft weighting scheme that emphasizes on harder examples.
In \cite{mentornet}, a student-teacher training framework is proposed such that the teacher model provides a curriculum via a sample weighting scheme for the student model to focus on samples whose labels are likely to be correct.
\cite{ghorbani2019data} studies the value of examples via Shapley values, and it shows that downweighting examples with low values might even improve performance.
Reweighing of examples is also used in self-paced learning \cite{kumar2010self,shu2019meta} where the weights are optimized to learn easier examples first. In \cite{wu2019adaframe}, an RL agent is used to adaptively sample relevant frames from videos.
In this paper, unlike the above, we focus on transfer learning -- L2TL formulates the transfer learning problem with a new loss function, including class-relevant weights and a dataset-relevant weights. L2TL learns the weight assignments with RL, in a setting where actions (source data selection) are guided with the rewards (target validation performance). Unlike gradient-descent based reweighing, RL-based rewarding is also applicable to scenarios where the target evaluation objective is non-differentiable, e.g., area under the curve (AUC).

\noindent\textbf{Meta learning:} Meta-learning broadly refers to learning to learn frameworks \cite{Schmidhuber1997} whose goal is to improve the adaptation to a new task with the information extracted from other tasks. Meta learners are typically based on inspirations from known learning algorithms like gradient descent \cite{maml} or derived from black box neural networks \cite{mann}.
As the notable meta learning application, in few-shot learning \cite{maml,zhu2020lim}, the use of validation loss as a meta-objective has been explored \cite{ravi2016optimization}.
However, for optimization problems with non-differentiable objectives like neural architecture search, RL-based meta-learning is shown to be a promising approach \cite{nas_rl,pham2018efficient}.
RL-based optimization has successfully been applied to other applications with enormously-large search spaces, e.g. learning a data augmentation policy~\cite{cubuk2018autoaugment}.
The specific form of RL application in L2TL is novel -- it employs guidance on the source dataset information extraction with the reward from the target validation dataset performance. 
Different from many meta learning methods, e.g. those for few-shot learning, we consider a common real-world scenario where a very large-scale source dataset is integrated to extract information from. We do not employ any episodic training, hence L2TL is practically feasible to employ on very large-scale source datasets.

\section{Learning From Source and Target Datasets}
We consider a general-form training objective function $\mathcal{L}(\mathbf{\Omega}, \mathbf{\zeta_S}, \mathbf{\zeta_T}, \lambda, \alpha_s, \alpha_t)$\footnote{Function arguments are not often shown in the paper for notational convenience.} jointly defined on a source dataset $D_S$ and a target dataset $D_T$:
\begin{equation}
    \begin{aligned}
    \mathcal{L} &= \alpha_s[i] \cdot \sum_{j = 1}^{B_S} \lambda(x_j, y_j; \mathbf{\Phi}) \cdot L_S(f_S(x_j; \mathbf{\Omega}, \mathbf{\zeta_S}), y_j)  \\
    &+ \alpha_t[i] \cdot \sum_{k=1}^{B_T} L_T(f_T(x_k^\prime; \mathbf{\Omega}, \mathbf{\zeta_T}), y_k^\prime), 
\label{optimization}
    \end{aligned}
\end{equation}
where $(x, y)$ are the input and output pairs ($x_j, y_j \sim D_S, x^\prime_k, y^\prime_k \sim D_T$),
$B_S$ and $B_T$ are the source and target batch sizes\footnote{Batch approximations may be optimal for different batch sizes for source and target dataset and thus may employ different batch normalization parametrization.},
$\alpha_s[i]$ and $\alpha_t[i]$ are the scaling coefficients at $i^{th}$ iteration,
$\lambda$ is the importance weighing function, $f_S(\cdot; \mathbf{\Omega}, \mathbf{\zeta_S})$ and $f_T(\cdot; \mathbf{\Omega}, \mathbf{\zeta_T})$ are encoding functions for the source and the target datasets with trainable parameters $\mathbf{\Omega}$, $\mathbf{\zeta_S}$ and $\mathbf{\zeta_T}$\footnote{In $f(\cdot;\mathbf{W})$ representation, $\mathbf{W}$ denote the trainable parameters.}.

To maximally benefit from the source dataset, a vast majority of the trainable parameters should be shared.
If we consider the decompositions, $f_S(\cdot; \mathbf{\Omega}, \mathbf{\zeta_S}) = h_S(\cdot; \mathbf{\zeta_S}) \circ g(\cdot; \mathbf{\Omega})$ and $f_T(\cdot; \mathbf{\Omega}, \mathbf{\zeta_T}) = h_T(\cdot; \mathbf{\zeta_T}) \circ g(\cdot; \mathbf{\Omega})$, $g$ shall be a high capacity encoder with large number of trainable parameters that can be represented with a deep neural network, and $h_T$ and $h_S$ are low capacity mapping functions with small number of parameters that can be represented with very shallow neural networks.\footnote{Source datasets are typically much larger and contain more classes, hence $h_S$ may have higher number of parameters than $h_T$.}
The learning goal of Eq. \ref{optimization} is generalizing to unseen target validation dataset, via maximization of the performance metric $R$:
\begin{equation}
    \sum_{x^\prime, y^\prime \sim D_T^\prime} R(f_T(x^\prime; \hat{\mathbf{\Omega}}, \hat{\mathbf{\zeta_T}}), y')). 
\end{equation}
$R$ does not have be differentiable with respect to $x$ and $y$ and may include metrics like the top-1 accuracy or area under the curve (AUC) for classification.
$\hat{\mathbf{\Omega}}$ and  $\hat{\mathbf{\zeta_T}}$ are the pre-trained weights optimized in Eq.~\ref{optimization}.

Without transfer learning, i.e., training with only target dataset, $\alpha_s[i]=0$ and $\alpha_t[i]=1$ for all $i$.
In fine-tuning, the optimization is first considered for the source dataset for $N_S$ steps with uniform weighing of the samples $\lambda(x, y)=1$, and then for the target dataset using the pre-trained weights $\hat{\mathbf{\Omega}}$, $\hat{\mathbf{\zeta_T}}$, i.e.:
\begin{equation}
(\alpha_s[i], \alpha_t[i])=\left\{\begin{matrix}
(1, 0), i < N_S, \\  
(0, 1), i > N_S.
\end{matrix}\right.
\label{alpha_schedule}
\end{equation}

Next, we describe our framework towards optimal learning from source and target datasets.

\section{Learning to Transfer Learn Framework}

\begin{figure*}[t]
\centering
\includegraphics[width=0.95\textwidth]{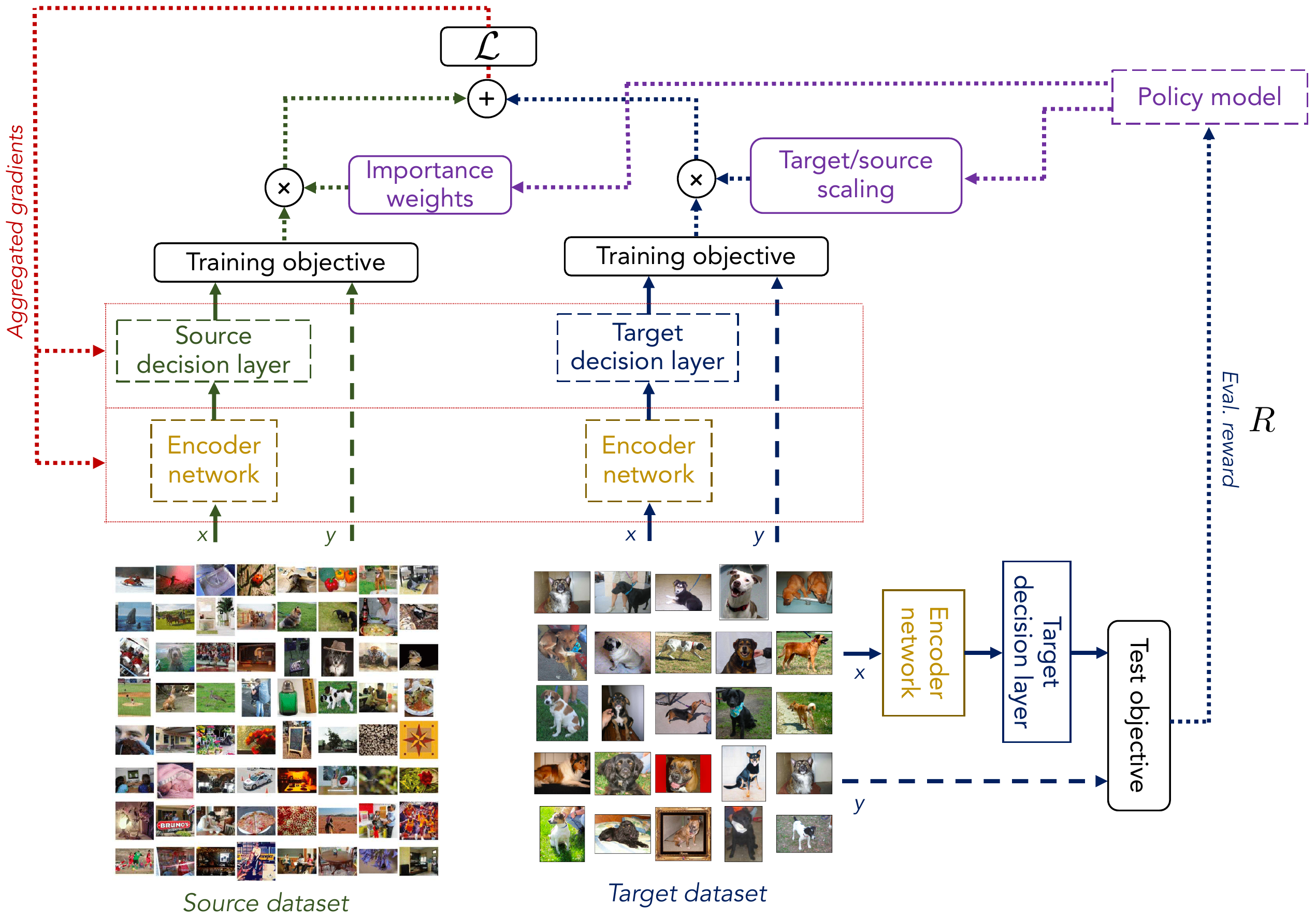}
\caption{{Overall diagram of the L2TL framework}. Dashed boxes correspond to trainable functions. L2TL employs a policy model to determine weigths of the source dataset samples, to extract the information in a careful way to maximize the target dataset test objective. The models on source and target datasets are shared, via the encoder network.}
\label{fig:overall}
\end{figure*}

We propose learning to transfer learn (L2TL) framework (shown in Fig. \ref{fig:overall}) to learn the weight assignment adaptively, rather than using a fixed weight assignment function $\lambda(x, y; \mathbf{\Phi})$ to measure the relatedness between the source domain and the target domain. Learning of the adaptive weights in L2TL is guided by the performance metric $R$ on a held-out target validation dataset. Thus, beyond targeting general relatedness, the framework directly targets relatedness for the specific goal of improvement in target evaluation performance.

While optimizing for $\lambda(x, y; \mathbf{\Phi})$, one straightforward option for scaling coefficients would be alternating them between $(1, 0)$ and $(0, 1)$ -- i.e. training the source dataset until convergence with optimized $\hat{\mathbf{\Phi}}$ and then training the target dataset until convergence with the pre-trained weights from the source dataset.
Yet, the approach may potentially require many alternating update steps and the computational cost may become prohibitively high.
Instead, we design the policy model in L2TL to output $(\alpha_s[i], \alpha_t[i])$ along with $\lambda$.\footnote{Without loss of generality, we can optimize a single weight ${\alpha_s[i]}$ (setting $\alpha_t[i]=1$) as the optimization is scale invariant.}
The policy optimization step is decoupled from the gradient-descent based optimization for $\mathbf{\Omega}$, $\mathbf{\zeta_S}$ and $\mathbf{\zeta_T}$. Updates are reflected to the policy model via the information embodied in $\mathbf{\Omega}$ and $\mathbf{\zeta_T}$. Algorithm.~\ref{algo:l2tl_algo} overviews the training updates steps.

\begin{algorithm}[t]
\DontPrintSemicolon
%\hrulefill
%\begin{algorithmic}
$N$ $\gets$ number of training iterations \\
\For {$i \gets 1$ to $N$} {
    $l_s \gets 0, l_t \gets 0$ \\
    \For {$j \gets 1$ to $B_S$} {
        Sample $x_j, y_j$ from $D_S$ \\
        Calculate classification loss $L_S(x_j, y_j; \mathbf{\Omega}, \mathbf{\zeta_S})$ \\
        Calculate example weight $\lambda(x_j, y_j; \mathbf{\Phi})$ \\
        $l_s = l_s + \lambda \cdot L_S $ \\
    }
    $l_s = \alpha_i \cdot l_s$ \\
    \For {$k \gets 1$ to $B_T$} {
        Sample $x^{\prime}_k$, $y^{\prime}_k$ from $D_T$ \\
        Calculate classification loss $L_T(x^{\prime}_k, y^{\prime}_k; \mathbf{\Omega}, \mathbf{\zeta_T})$ \\
        $l_t = l_t + L_T $ \\
    }
    Update $\mathbf{\Omega}, \mathbf{\zeta_S}, \mathbf{\zeta_T}$ using stochastic gradient descent with loss $l_s+l_t$\\
    $r \gets 0$ \\
    \For {$k \gets 1$ to $B_P$} {
        Sample $x^{\prime}_k$, $y^{\prime}_k$ from $D_{T'}$ \\
        Calculate reward $R(f_T(x^{\prime}_k; \mathbf{\Omega}, \mathbf{\zeta_T}), y^{\prime}_k)$ \\
        $r = r + R$ \\
    }
    Update $\mathbf{\Phi}$ with reward $r$ using policy gradient
}
%\end{algorithmic}
\caption{{\sc L2TL} -- Learning to Transfer Learn}
\label{algo:l2tl_algo}
\end{algorithm}

In the first phase of a learning iteration, we apply gradient decent-based optimization to learn the encoder weights $\mathbf{\Omega}$, and the classifier layer weights $\mathbf{\zeta_S}, \mathbf{\zeta_T}$ to minimize the loss function  $\mathcal{L}$:
\begin{equation}
\hat{\mathbf{\Omega}}, \hat{\mathbf{\zeta_S}}, \hat{\mathbf{\zeta_T}} = \text{argmin}_{\mathbf{\Omega}, \mathbf{\zeta_S}, \mathbf{\zeta_T}}\mathcal{L(\hat{\mathbf{\Phi}};\mathbf{\Omega}, \mathbf{\zeta_S}, \mathbf{\zeta_T})}.
\end{equation}
In this phase, the policy model is fixed, and its actions are sampled to determine weights.
Although most batches would contain relevant source dataset samples, the loss might be skewed if most of source dataset samples in a batch are irrelevant (and would ideally get lower weights). To ease this problem, we use a larger batch size and dynamically select the most relevant examples. At each iteration, we sample a training batch of size $M_S \cdot B_S$, and use the top $B_S$ of them with the highest weights for training updates. This approach also yields computational benefits as the gradients would not be computed for most source dataset samples until convergence.

In the second phase of a learning iteration, given encoder weights from the first phase, our goal is to optimize policy weights $\mathbf{\Phi}$ and maximize the evaluation metric $R_{D_T'}$ on the target validation set:
\begin{equation}
\max_{\mathbf{\Phi}} R_{D_T'}(\hat{\mathbf{\Omega}}, \hat{\mathbf{\zeta_S}}, \hat{\mathbf{\zeta_T}};\Phi).
\end{equation}
$D_T'$ is the held-out dataset to compute the reward. 
We treat this phase as an RL problem, such that the policy model outputs the action of value assignment for 
$\lambda(x, y; \mathbf{\Phi})$ and $\alpha$ towards optimization of the reward (with the environment being training and evaluation setup, and the state being the encoder weights as the consequence of the first phase).
In its general form $\lambda(x, y; \mathbf{\Phi})$ may yield a very high dimensionality for optimization of $\mathbf{\Phi}$.
For simplicity and computational-efficiency, we consider sample-independent modeling of $\lambda(x, y; \mathbf{\Phi})$, similar to \cite{ngiam2018domain},
i.e., $\lambda(x, y; \mathbf{\Phi}) =  \lambda(y; \mathbf{\Phi})$.\footnote{A search space with a higher optimization granularity is expected to improve the results, albeit accompanied by significantly increased computational complexity for meta learning of $x$-dependent $\lambda(x, y; \mathbf{\Phi})$.}

For more efficient optimization via efficient systematic exploration of the very large action space, we discretize the possible values of $\lambda(y; \mathbf{\Phi})$ into pre-defined number of actions, in the range $\lambda(y) \in [0, 1]$. We define $n$ actions, such that each action $k \in [0, n-1]$ corresponds to the weight value ${k}/{(n-1)}$. For example, when $n=11$, the weight values are $[0, 0.1, 0.2, \ldots, 1.0]$. We also discretize the possible values for $\alpha$, using $n'$ actions. Each action $k'$ corresponds to ${\beta k'}/{(n'-1)}$, where $\beta$ is a hyperparameter to constrain the value range of $\alpha$.
The search space has $n'\times(c_S)^{n}$ possibilities, where $c_S$ is the number of classes in the source dataset.
When training the policy model, we use policy gradient to maximize the reward on the target dataset $D_{T'}$, using a batch of $B_P$ samples. 
%We use the evaluation mode for the convolutional encoder.
At iteration $t$, we denote the advantage $A_t=R_t-b_t$, where $b_t$ is the baseline. 
Following~\cite{pham2018efficient}, we use the moving average baseline to reduce variance, i.e., $b_t=(1-\gamma)b_t+\gamma R_t$, where $\gamma$ is the decay rate.
The policy gradient is computed using REINFORCE~\cite{williams1992simple} and optimized using Adam~\cite{kingma2014adam}.

\section{Experiments}
\subsection{Datasets and Implementation Details}
We demonstrate the performance of L2TL in various scenarios.
As the source dataset, we use the ImageNet dataset \cite{russakovsky2015imagenet} containing 1.28M images from 1K classes, and also a much larger dataset, i.e., JFT-300M, containing $\sim$300M images from 18,291 classes to demonstrate the scalability of our approach. 
As the target datasets, we evaluate on \textbf{five} fine-grained image datasets (summarized in Table~\ref{table:dataset}): \textbf{Birdsnap}~\cite{berg2014birdsnap}, \textbf{Oxford-IIIT Pets}~\cite{parkhi2012cats}, \textbf{Stanford Cars}~\cite{krause2013collecting}, \textbf{FGVC Aircraft}~\cite{maji2013fine}, and \textbf{Food-101}~\cite{bossard2014food}. In addition, we also consider transfer learning scenario from MNIST to SVHN~\cite{netzer2011reading} to assess the effectiveness of L2TL for small-scale source datasets.

%Top-1 accuracy is reported for all the experiments.
We also consider two target datasets with classes that do not exist in the source datasets: Describable Textures Dataset (DTD)~\cite{cimpoi14describing} and Chest X-Ray Dataset CheXpert~\cite{irvin2019chexpert}.
\noindent\textbf{Describable Textures Dataset:} DTD contains textural images in the wild from 47 classes such as striped and matted.
The dataset has 20 splits and we evaluate the testing results on the first split.
Each training, validation, and testing split has 1,880 images. 
\noindent\textbf{Chest X-Ray Dataset:} The CheXpert medical dataset is used for chest radiograph interpretation task. It consists of 224,316 chest radiographs of 65,240 patients labeled for 14 observations as positive, negative, or uncertain.
Following~\cite{irvin2019chexpert}, we report AUC on five classes and we regard ``uncertain'' examples as positive.\footnote{Our reproduced results are matched with~\cite{irvin2019chexpert} on mean AUC. However, there are variances as we can see that for some classes, we achieve slightly worse than~\cite{irvin2019chexpert}. This may because of the small number of validation examples (200) used.} For L2TL, we use the mean AUC as the reward.

\noindent\textbf{Implementation Details.}
When the source dataset is ImageNet, we use a batch size $B_S=256$, $B_T=256$, $B_P=1024$ and a batch multiplier $M_S=5$ for all the experiments.
For the JFT-300M dataset, to reduce the number of training iterations, we use $B_S=1,024$.
The number of actions $n^\prime$ for $\alpha$ is 100.

We use Inception-V3 for all the experiments except CheXpert.
For target dataset, we search the initial learning rate from $\{0.001$, $0.005$, $0.01$, $0.05$, $0.1$, $0.15$, $0.2$, $0.4\}$, and weight decay from $\{0, 4\times10^{-5}\}$.
All the datasets are optimized using SGD with a momentum of 0.9, trained for 20,000 iterations. We use the single central crop during evaluation.
The learning rate is cosine decayed after first 2,000 iterations warmup.
When optimizing our policy model, we use the Adam optimizer with a fixed learning rate 0.0001. As policy model parameters, we set $\beta=0.5$ and $\gamma=0.05$.
We follow the standard image preprocessing procedure for Inception-V3 on both the source images and the target images.

\begin{table}[t]
\centering
\caption{{Details of the five fine-grained datasets}:
Birdsnap (Birds)~\cite{berg2014birdsnap}, Oxford-IIIT Pets (Pets)~\cite{parkhi2012cats}, Stanford Cars (Cars)~\cite{krause2013collecting}, FGVC Aircraft (Air)~\cite{maji2013fine}, and Food-101 (Food)~\cite{bossard2014food}.
}
\label{table:dataset}
\begin{tabular}{l|c|c|c|c|c} 
\hline
                & Birdsnap & Oxford-IIIT Pets & Stanford Cars &  Aircraft  & Food-101 \\  
\hline
\# of classes   &  500     &   37             & 196           &  100          &  101      \\
\hline
\# of train examples  & 42,405   & 2,940            & 6,494         &  3,334        & 68,175      \\
\hline  
\# of valid examples  &  4,981   & 740              & 1,650         &  3,333        & 7,575 \\
\hline 
\# of test examples   & 2,443    & 3,669            & 8,041         &  3,333        & 25,250 \\
\hline
\end{tabular} 
\end{table}

\begin{table}[t]
\centering
%\begin{tabular}{|C{1.9 cm}|C{.8 cm}|C{.8 cm}|C{.8 cm}|C{1.0 cm}|C{.8 cm}|}
\caption{{Transfer learning performance with ImageNet source dataset}. * indicates our implementation.
}
\label{table:transfer_results}
  \setlength{\tabcolsep}{1.5mm}
{
\begin{tabular}{c|c|c|c|c|c} 
\hline
  \multirow{2}{*}{Method} & \multicolumn{5}{|c}{Target dataset test accuracy (\%)} \\
 \cline{2-6}
  & Birdsnap & Oxford-IIIT Pets & Stanford Cars &  Aircraft  & Food-101 \\  
\hline
Fine-tuning \cite{ngiam2018domain} &  77.2 &  93.3 &  91.5   &  88.8   &  88.7     \\ 
 Fine-tuning*            &  77.1 &  93.1 &  92.0   &  88.2   &  88.4  \\ \hline
    MixDCNN~\cite{wang2015multiple} &    74.8       &   -    & - &   82.5  &  - \\
    EMD \cite{cui2018large}  &  -    &    -      &   91.3     &  85.5  &  88.7 \\
    OPAM \cite{peng2017object} &   -   &  93.8     &  92.2      &    -   &  -   \\
 DATL \cite{ngiam2018domain} &  76.6 &  94.1 &  92.1   &  87.8   &  88.9     \\ 
\hline
 Our L2TL                &  \textbf{78.1} &  \textbf{94.4} &  \textbf{92.6}   &  \textbf{89.1}   &  \textbf{89.2}   \\
\hline
\end{tabular} 
}
\end{table}

For CheXpert, we use the DenseNet-121 architecture~\cite{huang2017densely} and follow the evaluation protocol specified in~\cite{irvin2019chexpert}, where ten crops are used for evaluation and 30 checkpoints are ensembled to obtain the final results.
We cross validate weight decay and initial learning rate,
where the weight decay is searched in [0, 0.0001] and the learning rate searched in range $[0.5, 0.8, 1.0, 1.3, 1.5, 2.0]$.
All other hyperparameters are same as above.
We use the same input preprocessing as described in \url{https://github.com/zoogzog/chexnet}.

\begin{table}[t]
\centering
\caption{{Results on Birdsnap using JFT-300M as the source dataset}. The performance is reported on the test set.}
\label{table:bird_jft_res}
  \setlength{\tabcolsep}{5.0mm}
  {
\begin{tabular}{  l | c  }
\hline
 Method         &  Birdsnap accuracy (\%)    \\
\hline
 Fine-tuning      &   74.9            \\
\hline
 DATL \cite{ngiam2018domain}           &   81.7           \\
\hline
 Our L2TL           &   \textbf{82.4}           \\
\hline
\end{tabular} 
}
\end{table}

\begin{table}[t]
\centering
\caption{{Transfer learning from MNIST to SVHN.} As shown in \cite{liu2019understanding}, fine-tuning shows gains over other training cases due to the inherent similarity between datasets. L2TL efficiently exploits this further by emphasizing on some MNIST classes more than others, and improves the transfer learning gains significantly. 
}
\label{table:transfer_mnist}
  \setlength{\tabcolsep}{4.0mm}{
\begin{tabular}{  l | c  }
\hline
 Method         &  SVHN accuracy (\%)    \\
\hline
Random initialization     &      64.8          \\
\hline
 Fine-tuning      &   71.7           \\
\hline
 Our L2TL           &   \textbf{75.2}           \\
\hline
\end{tabular}
}
\end{table}

Hyperparameters of the encoder models are chosen from the published baselines and the policy model parameters are cross-validated on a validation set. 
For datasets that the testing accuracy is reported using the model trained on training and validation samples, L2TL is first trained on the training set using the reward from the validation set. Then, the learned control variables are used to train the joint model on the combined set of training and validation samples -- we completely exclude the test set during training. For the fine-tuning experiments, we use the best set of hyperparameters evaluated on the validation set.
We present the results averaged over three runs. We observe that the standard deviation for the L2TL accuracy to be around 0.1\%, much smaller than the gap between different methods. 

\subsection{Similar domain transfer learning}
\label{sec:label_overlap}

We initially consider the scenario of target datasets with classes that mostly exist in the source dataset.

\begin{figure}[t]
\centering
\includegraphics[width=1.0\textwidth]{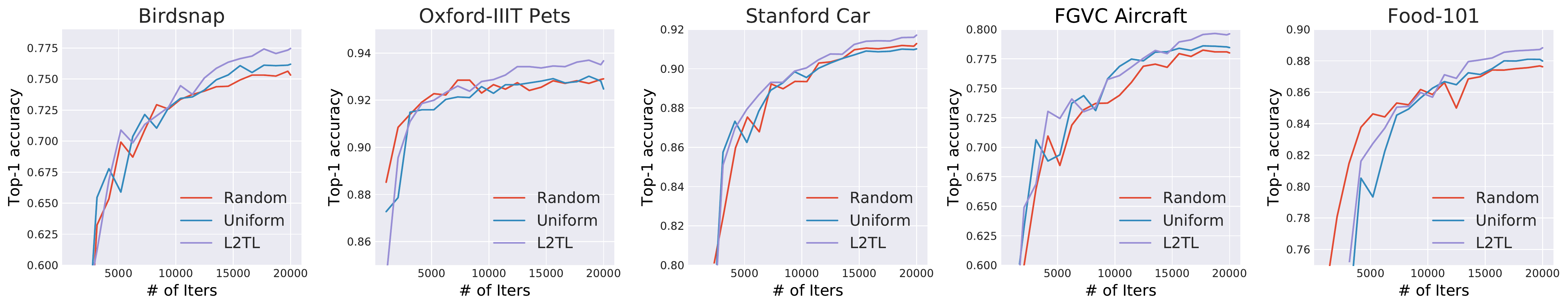}
\caption{{Performance comparisons between L2TL, random search and uniform weights}. The curves are oscillatory at the beginning, but become stable later during the training. L2TL outperforms the baselines when the training converges.}
\label{fig:random_vs_rl}
\end{figure}

\begin{figure}[t]
\centering
\includegraphics[width=1.0\textwidth]{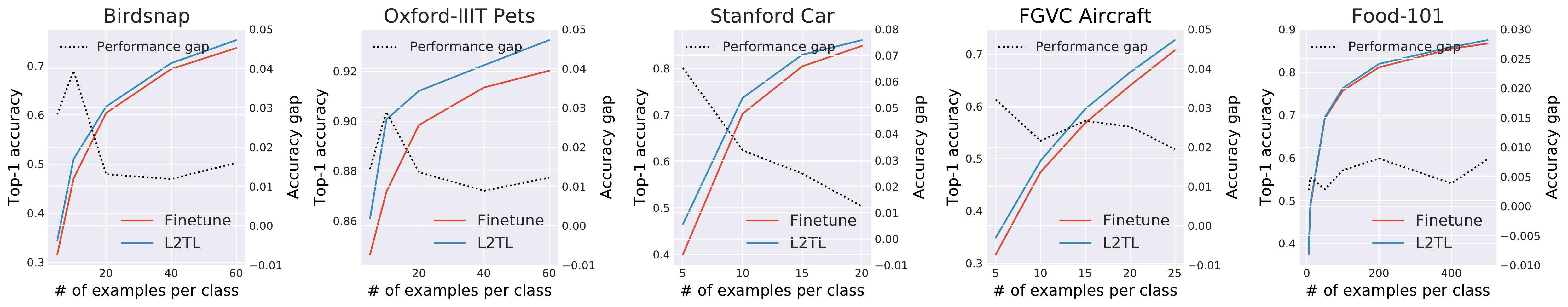}
\caption{{Number of examples per class vs. top-1 accuracy for L2TL and fine-tuning}.}
\label{fig:low_shot_results}
\end{figure}

\begin{table}[t]
\centering
\caption{{Target dataset test accuracy (\%) on Stanford Car with different number of target samples per class}.
}
\label{table:cars_small_data}
\setlength{\tabcolsep}{3.0mm}{
\begin{tabular}{c|c|c|c|c} 
\hline
\multirow{2}{*}{Method} & \multicolumn{4}{|c}{ Number of samples per class} \\
 \cline{2-5}
  & 5   & 10  &  15  &  20    \\   \hline

 Fine-tuning            &  40.0 &  70.3 &  80.5   &  84.9 \\
 \hline
 Our L2TL                &  \textbf{46.5} &  \textbf{73.7} &  \textbf{83.0}   &  \textbf{86.1}   \\
\hline
\end{tabular} 
}
\end{table}

\begin{figure}[t]
\centering
\includegraphics[width=1.0\textwidth]{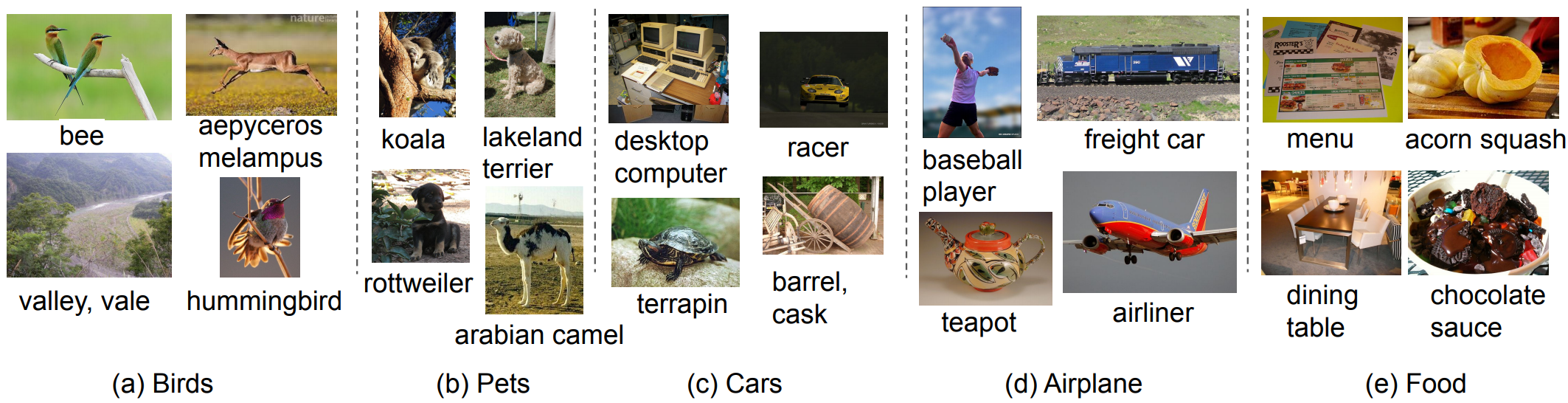}
\caption{{Representative examples from the source datasets with high weight from L2TL for different target datasets}. In most cases, we observe the selected examples to be highly-related to the target dataset.}
\label{fig:related_cats}
\end{figure}

\noindent\textbf{Performance and comparison to other transfer learning methods.}
We first evaluate L2TL on five fine-grained datasets focusing on different subsets, with the reward of validation set top-1 accuracy.
Table~\ref{table:transfer_results} shows the results of L2TL along with fine-tuning and state-of-the-art transfer learning benchmarks.
With a well-optimized network architecture and learning rate scheduling, fine-tuning is already a solid baseline for the datasets in Table~\ref{table:dataset} \cite{cui2018large,ngiam2018domain}.
Yet, L2TL outperforms fine-tuning across all the datasets with 0.6\%-1.3\% accuracy difference, which demonstrates the strength of L2TL in selecting related source examples across various domains. DATL performs worse than fine-tuning on Birdsnap and Aircraft, unlike L2TL. This underlines the importance of leveraging the visual similarity in the ways beyond label matching as in DATL. 
When the much larger JFT-300M source dataset is considered, Table~\ref{table:bird_jft_res} shows that L2TL shows even greater benefits in learning related samples to extract knowledge from despite the large-scale of options, demonstrating \textbf{7.5\%} improvement over the fine-tuning baseline.

\begin{figure}[t]
\centering
\includegraphics[width=1.0\textwidth]{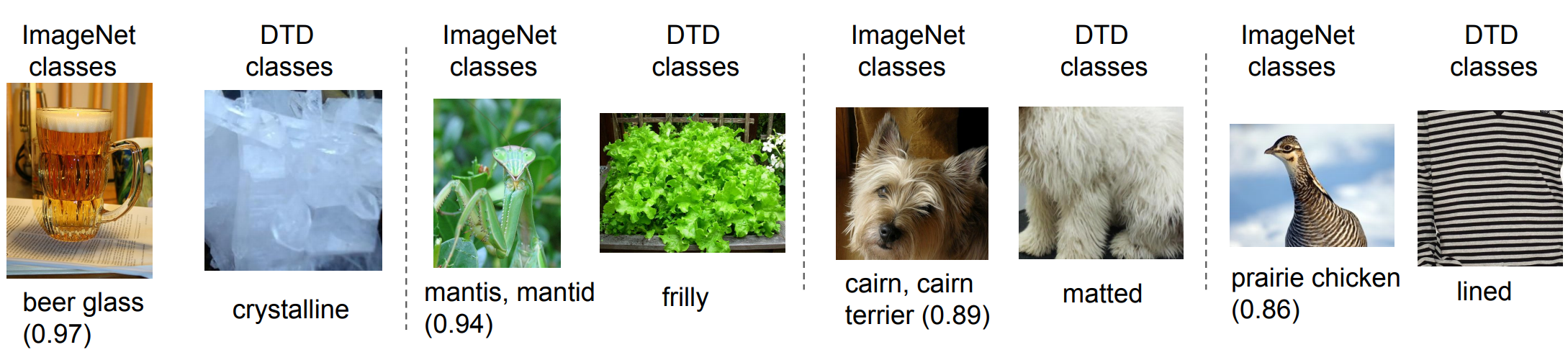}
\caption{{Top source classes with the highest weight from ImageNet while transferring to DTD target}. Representative images from each ImageNet class are shown along with related examples from DTD.}
\label{fig:related_dtd}
\end{figure}

\begin{table}[t]
\centering
\caption{{Results on the test set on DTD, split 1}.}
\label{table:dtd_res}
  \setlength{\tabcolsep}{2.5mm}{
\begin{tabular}{ l|  l | c  }
\hline
 Number of Training examples & Method         &  Accuracy (\%) \\ 
 \hline
 & Random initialization      &  57.4 \\ 
Full training set & Fine-tuning      &   70.3            \\
 & L2TL           &   \textbf{72.0}           \\
\hline\hline
\multirow{2}{*}{10 examples per category} & Fine-tuning      &      55.0 \\ 
 & Our L2TL           &     \textbf{60.1}      \\
\hline
\end{tabular} 
}
\end{table}

We additionally conduct the experiments on transfer learning from MNIST to SVHN~\cite{netzer2011reading}, in a setting similar to \cite{liu2019understanding}. Although both datasets correspond to the same content of digits, the font style of digits are quite distinct, with varying degree of differences among individual digits.
Following \cite{liu2019understanding}, we construct the SVHN dataset by randomly sampling 60 images per class, resulting in 600 images in the training split. We use the pretrained LeNet~\cite{lecun1998gradient} for transferring source dataset of MNIST to the target dataset of SVHN. Table~\ref{table:transfer_mnist} shows that comparing to fine-tuning, L2TL obtains more than {3.5\%} improvement in performance, via upweighing of the relevant digit images from MNIST. This also validates the effectiveness of L2TL even with small-scale source datasets.

\begin{table}[t]
\centering
\caption{{AUC comparisons on the CheXpert dataset}. We followed the the same evaluation protocol in~\cite{irvin2019chexpert} .
}
\label{table:chest_xray}
\begin{adjustbox}{width=1.0\textwidth}
\begin{tabular}{  l | c | c | c | c | c | c  }
\hline
Method                                & Atelectasis  & Cardiomegaly  & Consolidation & Edema & Pleural Effusion  & Mean \\
\hline
Fine-tuning~\cite{irvin2019chexpert}                     &  85.8        &   83.2        &  89.9         &  94.1          &    \textbf{93.4}   & 89.3       \\
\hline                     
Fine-tuning            &  85.2        &   83.8        &  90.0         &  94.5          &   92.8    & 89.3       \\
\hline            
   Our L2TL                        &\textbf{86.1} &  \textbf{84.4}&\textbf{91.5}  & \textbf{94.8}  &   93.2    & \textbf{90.0}        \\
\hline
\end{tabular} 
\end{adjustbox}
\end{table}

\noindent\textbf{Learning importance weights.}
We study the effectiveness of learning importance weights in L2TL by comparing to two baselines:
(i) random search: the policy model is not optimized and random actions are chosen as the policy output, and (ii) uniform weights: a constant importance weight is assigned to all training samples. Note that for these baselines, $\alpha$ is still optimized via policy gradient.
We show the best results of the baselines, after optimizing the hyperparameters on the validation set.
As shown in Fig.~\ref{fig:random_vs_rl}, L2TL outperforms both after sufficient number of iterations, demonstrating the importance of reweighting via policy gradient.
L2TL converges to the final result in the last few thousand iterations, although a larger variance is observed at the beginning of training.
We do not observe large variations in the final performance with different %experimental
runs (e.g. the standard deviation of performance is around 0.1 \% over 3 runs).
As the classifiers of both the source and the target dataset converge, the small variations in the weights for each class would not heavily affect the final performance. 

\noindent\textbf{Small target dataset regime.}
In the extreme regime of very small number of training examples, generalizing to unseen examples is particularly challenging as the model can be prone to overfitting. Fig. \ref{fig:low_shot_results} shows that in most cases, we observe significant increase in performance when the number of examples per class is smaller.
For five examples per class, the gap is as high as \textbf{6.5\%} (for Stanford Car) (see Table \ref{table:cars_small_data}).
We observe that the gap between the L2TL and the fine-tune baseline often becomes smaller when more examples are used, but still remains as high as 1.5\% with 60 examples per class (for Birdsnap). These underline the potential of L2TL for significant performance improvements in real-world tasks where the number of training examples are limited.

\noindent\textbf{High-weight source samples.}
To build insights on the learned weights, we sample 10$k$ actions from the policy and rank the source labels according to their weights. 
For Birds, the top source class is ``bee eater'' which is one of the bird species in ImageNet. The second top ``aepyceros melampus'' is an antelope that has narrow mouth, which is similar to some birds with sharp spout. The ``valley'' also matches the background in some images. For Cars, we interestingly observe the high-weight class ``barrel, cask'', which indeed include wheels and car-looking body types in many images. ``Terrapin'' is a reptile that crawls on the ground with four legs, whose shape looks like vehicles in some way.
For Food, the high-weight classes seem relevant in a more subtle way -- e.g., ``caldron'' might have images with food inside.
%, but most ``seashore, coast'' are less related to food.
Fig.~\ref{fig:related_cats} visualizes a few representative examples from each dataset. More classes can be found in our supplementary material.
These demonstrate that L2TL can carefully extract the related classes from the source based on the pattern/shape of the objects, or background scenes.
L2TL yields ranking of the source data samples, which can be utilized as new forms of interpretable insights for model developers.

\subsection{Dissimilar domain transfer learning}
We evaluate L2TL on datasets that are dissimilar to the source dataset, where alternative methods like DATL cannot be readily applied.
Table~\ref{table:dtd_res} shows the results on DTD. We observe that ImageNet fine-tuning greatly improves the classification results compared to training from random initialization. L2TL further improves the fine-tuning baseline by 1.5\%, demonstrating the strength of L2TL selectively using related source classes instead of all classes. 
For the low-shot target dataset regime, with 10 examples per class, the improvement is more than 5\%, suggesting the premise of L2TL even more strongly.
Fig.~\ref{fig:related_dtd} shows that L2TL is able to utilize visually-similar patterns between the source and the target classes. The similarities occur in the form of texture pattern for most DTD classes. For example, ``praire chicken'' images from ImageNet typically contain patterns very relevant to ``lined'' from DTD. Training with such visually-similar patterns especially helps the low layers of the networks as they can reuse most of the relevant representations when transferring knowledge \cite{liu2019understanding}.

Similarly, Table~\ref{table:chest_xray} shows the results on CheXpert, using target validation AUC as a L2TL reward. L2TL performs better than the fine-tuning baseline with an AUC improvement of 0.7. There are not many straightforward visual similarities to humans between ImageNet and CheXpert, but L2TL is still capable of discovering them to improve performance.

\begin{table}[t]
\centering
\caption{{Computational cost of training on Cloud TPU v2. We use Inception-V3 as the backbone}. 
The last column (``With PT model'') assumes availability of a pre-trained source model. ``TL'' denotes transfer learning from source to target.
}
\label{table:training_time}
\begin{tabular}{c|c|c|c|c|c|c}
\hline
\multirow{2}{*}{Method} & \multicolumn{2}{|c|}{Number of iterations} & \multicolumn{2}{|c|}{Time per iterations} & \multicolumn{2}{|c}{Total time} \\
\cline{2-7}
             & Pre-training      & TL  & Pre-training    & TL & 
             From scratch & With PT model\\
\cline{1-7}
Fine-tuning  &  213,000          &   20,000  &  0.14s          &  0.21s         & 9.5h & 1.2h  \\
\cline{1-7}             
DATL  \cite{ngiam2018domain}       &  713,000 &   20,000  &  0.14s          &  0.21s         & 28.9h & 20.6h \\
\cline{1-7}             
Our L2TL         & 213,000           & 20,000    &  0.14s          &  0.75s         & 12.5h & 4.2h \\
\hline
\end{tabular} 
\end{table}

\subsection{Computational cost of training}
\textit{L2TL uses both the source and target data for training, and the source data can be potentially very large, but the excess computational overhead of L2TL is indeed not large}. 
Table~\ref{table:training_time} presents the computational cost for fine-tuning, DATL and L2TL with Imagenet as the source dataset.
In DATL, given a new target dataset, a new model has to be trained on the resampled data until convergence.
This step is time-consuming for large-scale source datasets. 
In L2TL, the transfer learning step is more expensive than fine-tuning, as it requires the computation on both source and target datasets. 
Yet, it only requires a single training pass on the source dataset, and thus the training time is much lower compared to DATL, and only $\sim30\%$ higher than fine-tuning when the whole training is considered. 

\section{Conclusions}

We propose a novel RL-based framework, L2TL, to improve transfer learning on a target dataset by careful extraction of information from a source dataset. 
We demonstrate the effectiveness of L2TL for various cases. L2TL consistently improves fine-tuning across all datasets. The performance benefit of L2TL is more significant for small-scale target datasets or large-scale source datasets. Even for the cases where source and target datasets come from substantially-different domains, L2TL still yields clear improvements.

\clearpage
\bibliographystyle{splncs04}
\bibliography{egbib}

\clearpage
\appendix

\section{High-weight classes from ImageNet}

\begin{table}[!htb]
\centering
\caption{{Top chosen classes from ImageNet source dataset}.}
\label{table:class scores}
\begin{tabular}{ l | c  }
\hline
Target dataset & Source classes (weights) \\
\hline
     
\hline
      & bee eater (1.0); aepyceros melampus (0.95); sea cradle (0.92); \\ 
Birds & barracouta (0.91); valley (0.90); sombrero (0.86); rosehip (0.84);  \\
      & Scottish deerhound (0.82); black swan (0.77); bell pepper (0.77); \\
      %coyote (0.77); American robin (0.76); \\
\hline
     & coral reef (0.88); prayer rug (0.88); koala (0.84); fire salamander (0.81) \\
Pets     & Irish setter (0.79); Arabian camel (0.78); Irish terrier (0.74);\\
         &  leaf beetle (0.72);  Rottweiler (0.71); giant schnauzer (0.70);\\
         % English setter (0.69) \\
\hline
     & desktop computer (0.80); butternut squash (0.76);  barrel, cask (0.65); \\
Cars & weevil (0.60); pool table (0.56); clumber (0.54); passenger car (0.50); \\
     & race car (0.49); washer (0.46); terrapin (0.45); seat belt (0.32); \\ % fire truck (0.29); \\
\hline
         &  bagel, beigel (0.97); ballplayer, baseball player (0.84); freight car (0.80); \\
Aircraft & teapot (0.83); crate (0.78); velvet (0.74); electric locomotive (0.68); \\
         & pirate, pirate ship (0.45); amphibious vehicle (0.28); airliner (0.10);\\
\hline
         & caldron, cauldron (0.84); menu (0.81); seashore, coast, seacoast (0.74) \\
Food     & acorn squash (0.73); dining table, board (0.67); globe artichoke (0.67); \\
         & mushroom (0.65); chocolate sauce, chocolate syrup (0.62); plate (0.54); \\ % sea urchin (0.5374) \\
\hline
\end{tabular}
%\caption{\textbf{Top chosen classes from ImageNet source dataset that are related to the target datasets}.}
\end{table}

For the target datasets, i.e., Birdsnap, Oxford-IIIT Pets, Stanford Cars, FGVC Aircraft, Food-101,
we list the high-weight classes from the source ImageNet in Table~\ref{table:class scores}. These classes are most related to the target dataset.
Some top classes do not from the same categories, but they are visually similar.
For instance, the top related class for ``Stanford Cars'' is ``desktop computer'', where the ``desktop computer'' class like boxes.
The ``valley'' class matches the background in some examples in the Birdsnap dataset.

For the Aircraft dataset, the ``teapot'' class is assigned with weight 0.83. We observe that the spout of teapot and teapot handle resemble the wing of an airplane. This indicates that our model is able to learn features that are implicitly relevant to the target samples and would benefit the learning process, even though the features may belong to different classes. Our L2TL yields ranking of the source data samples, which can be utilized as new forms of interpretable insights for model developers.

\section{More details on REINFORCE}

We add more details about the REINFORCE algorithm. The objective is
$J(\mathbf{\Phi})=E_{\lambda\sim\pi_{\mathbf{\Phi}}}r(\lambda)$, where $\mathbf{\Phi}$ is the parameters to be optimized, $r(\lambda)$ is the reward, and $\lambda$ is a sampled action. We aim to optimize parameters by maximizing $J(\mathbf{\Phi})$. The gradient of $J(\mathbf{\Phi})$ is
\begin{equation}
    \nabla_{\mathbf{\Phi}}J(\mathbf{\Phi}) = E_{\lambda\sim\pi_{\mathbf{\Phi}}(\lambda)}\nabla_{\mathbf{\Phi}}\log \pi_{\mathbf{\Phi}}(\lambda))r(\lambda).
\end{equation}
Monte Carlo sampling is used to obtain $N$ sequences to approximate the policy gradients:
\begin{equation}
\nabla_{\mathbf{\Phi}}J(\mathbf{\Phi}) \approx \frac{1}{N}\sum_{i=1}^N\sum_{t=1}^{T}\nabla_{\mathbf{\Phi}}\log\pi_{\mathbf{\Phi}}({\lambda_{i,t}} ) r({{\lambda_{i,t}}}),
\end{equation}
where $T$ is the length of an episode.

\end{document}